\documentclass[acmlarge,screen]{acmart}
\AtBeginDocument{%
  \providecommand\BibTeX{{%
    \normalfont B\kern-0.5em{\scshape i\kern-0.25em b}\kern-0.8em\TeX}}}

\setcopyright{rightsretained}
\copyrightyear{2021}
\acmYear{2021}
\acmDOI{}
\acmISBN{} 

\acmConference[KDD '21]{KDD '21: ACM SIGKDD Conference on Knowledge Discovery and Data Mining}{August 14--18, 2021}{Singapore}
\acmBooktitle{ACM SIGKDD Conference on Knowledge Discovery and Data Mining, August 14--18, Singapore, 2021.}

\usepackage{soul}
\usepackage{color}

\iffalse
    
    \newcommand{\bk}[2]{\textcolor[RGB]{50,150,50}{#2}}
    \newcommand{\john}[2]{\textcolor[RGB]{50,150,50}{#2}}
    \newcommand{\daudt}[2]{\textcolor[RGB]{50,150,50}{#2}}
    
\else
    
    \newcommand{\bk}[2]{#2}
    \newcommand{\john}[2]{#2}
    \newcommand{\daudt}[2]{#2}
    
\fi

\begin{document}

\title{Mapping Vulnerable Populations with AI}


\author{Benjamin Kellenberger}
\authornote{Both authors contributed equally to this research.}
\email{benjamin.kellenberger@epfl.ch}
\orcid{0000-0002-2902-2014}
\author{John E. Vargas-Muñoz}
\email{john.vargas@epfl.ch}
\authornotemark[1]
\author{Devis Tuia}
\email{devis.tuia@epfl.ch}
\orcid{0000-0003-0374-2459}
\affiliation{%
  \institution{Ecole Polytechnique Fédérale de Lausanne (EPFL)}
  \streetaddress{Rue de l’Industrie 17}
  \city{Sion}
  \state{Valais}
  \country{Switzerland}
  \postcode{CH-1951}
}

\author{Rodrigo C. Daudt}
\orcid{0000-0002-4952-9736}
\author{Konrad Schindler}
\orcid{0000-0002-3172-9246}
\affiliation{%
  \institution{Photogrammetry and Remote Sensing, ETH Zurich}
  \streetaddress{8092}
  \city{Zurich}
  \country{Switzerland}}
\email{(e-mail)}

\author{Thao Ton-That Whelan}
\author{Brenda Ayo}
\affiliation{%
  \institution{International Committee of the Red Cross (ICRC)}
  \streetaddress{Avenue de la Paix, 19}
  \city{1202 Geneva}
  \country{Switzerland}}
\email{ttonthat@icrc.org}

\author{Ferda Ofli}
\email{fofli@hbku.edu.qa}
\author{Muhammad Imran}
\email{mimran@hbku.edu.qa}
\affiliation{%
  \institution{Qatar Computing Research Institute, HBKU}
  \streetaddress{HBKU Research Complex}
  \city{Doha}
  \country{Qatar}}


\renewcommand{\shortauthors}{Kellenberger et al.}

\begin{abstract}
    Humanitarian actions require accurate information to efficiently delegate support operations. Such information can be maps of building footprints, building functions, and population densities. While the access to this information is comparably easy in industrialized countries thanks to reliable census data and national geo-data infrastructures, this is not the case for developing countries, where that data is often incomplete or outdated. Building maps derived from remote sensing images may partially remedy this challenge in such countries, but are not always accurate due to different landscape configurations and lack of validation data. Even when they exist, building footprint layers usually do not reveal more fine-grained building properties, such as the number of stories or the building's function (e.g., office, residential, school, etc.). In this project we aim to automate building footprint and function mapping using heterogeneous data sources. In a first step, we intend to delineate buildings from satellite data, using deep learning models for semantic image segmentation. Building functions shall be retrieved by parsing social media data like for instance tweets, as well as ground-based imagery, to automatically identify different buildings functions and retrieve further information such as the number of building stories. Building maps augmented with those additional attributes make it possible to derive more accurate population density maps, needed to support the targeted provision of humanitarian aid.
\end{abstract}



\maketitle

\section{Introduction}

Population maps are important for humanitarian organizations to reach people in need and to allocate their resources efficiently (e.g., in disaster relief or vaccination campaigns). Obtaining up-to-date information about the population distribution is challenging for many countries of the global South, due to outdated censuses (sometimes by more than 10 years\footnote{\url{https://unstats.un.org/unsd/demographic-social/census/censusdates}})
unable to keep up with population growth, migration and urbanization~\cite{thomson2020gridded}. Moreover, even if censuses are held on a periodic basis, they typically do not provide any population \emph{distribution} maps and are therefore of limited use, e.g. for disaster response planning within a city.

To provide automated estimates of population distribution, several works have used building maps as a proxy, e.g.,~\cite{weber2018census}. However, despite the efforts of humanitarian mapping projects (e.g., HOT-OSM\footnote{\url{https://www.hotosm.org}} and Missing Maps\footnote{\url{https://www.missingmaps.org}}), building maps are still missing or incomplete in (parts of) many developing countries. And even if maps exist, they often lack information about the usage of the buildings. Such information is, however, crucial for field interventions, as the expected number of people differs between, say, a residential building, a hospital and warehouse of the same size.

Projects like the Facebook population density maps~\cite{tiecke2017mapping} have applied computer vision methods to remote sensing imagery to automate the creation of building maps for entire countries. However, these maps neither delineate individual buildings (but rather patch-level predictions) nor contain information about usage, height, etc.  

To complement remote sensing images, social media data also carry information about human use of space. Social media usage is widespread around the world and can provide insights about human settlements. For instance, recent works have studied Tweets as a source for demographic analysis, and advertisement data from Facebook as an information source about poverty~\cite{fatehkia2020relative}.

Here we introduce a research project that aims at mapping populations in cities, exploiting synergies between remote sensing and social media. With an interdisciplinary team of experts from the humanitarian sector and academia, we are developing methods to combine these heterogeneous data sources and produce reliable building maps and population density estimates at a fine-grained level. Our objective is to deliver actionable information for personnel on the ground during and after conflicts, natural disasters and other events that require humanitarian actions.

\section{Related Work}

Population mapping by processing remote sensing imagery has been studied in several projects. The projects Global Human Settlement Layer~\cite{pesaresi2013global} and Facebook population density maps~\cite{tiecke2017mapping} detect built-up areas via patch-wise building classification to estimate population densities, respectively numbers. However, no further building attributes are estimated, such that the step from (approximate) building area to population numbers introduces a large margin of error.

Most recent methods proposed to extract \bk{}{urban land cover or} buildings from remote sensing images~\cite{hamaguchi2018building, shi2020building, han2020lightweight} use deep learning techniques trained on large datasets to obtain per-pixel labels \emph{building} or \emph{no building}. While one could, in principle, obtain population estimates~\bk{}{\cite{han2020learning}, or even socioeconomic indicators~\cite{tingzon2019mapping},} directly in that raster format, it is usually transformed into a vector data to simplify its further use in Geographic Information Systems. Some works obtain vector outlines in post-processing, by polygonizing raster predictions~\cite{tasar2018polygonization}, whereas others learn to produce vector outputs end-to-end~\cite{marcos2018learning, cheng2019darnet}, sometimes supported by specific shape priors for, e.g., rural buildings~\cite{vargas2019correcting}.   

Building use classification is also an active research topic. Different authors have been considering remote sensing data~\cite{xie2017classification} ground-based pictures (e.g., Google Street View)~\cite{tracewski2017repurposing, srivastava2020fine}, and also hybrid approaches that fuse these two data sources~\cite{srivastava2019understanding, workman2017unified}. However, ground-based pictures are often unavailable, such that these fusion methods must be able to deal with missing data~\cite{srivastava2019understanding}. Beyond image data,~\cite{haberle2019geo} explores the use of geo-located tweet messages as input to distinguish coarse use classes (i.e., residential and commercial buildings). In a similar spirit, advertisement data from Facebook has been used to estimate wealth indices for clusters of households~\cite{fatehkia2020relative}, but not at the building level.

Regarding building height, different approaches have been proposed using stereo imaging~\cite{Facciolo2017,Stucker2020}, Synthetic Aperture Radar (SAR)~\cite{sun2017building} or Light Detection and Ranging (LiDAR)~\cite{wu2019city}. Height estimation using stereo image pairs or sets requires high resolution images, which may be prohibitively expensive at larger scales. Furthermore, the on-demand acquisition of such images can quickly become affected by weather conditions.
To allow for fine-grained estimates at individual building level from SAR, high resolution and often multiple images of the same region are required~\cite{sun2019large}.
It has thus been proposed~\cite{srivastava2017joint} to jointly estimate terrain elevation and land use from monocular, optical remote sensing imagery, with a multi-task framework. 
Generally, topographical variations tend to degrade the height estimation from satellite imagery, and the application in hilly or mountainous terrain remains an open problem.
LiDAR with sufficiently small footprint (respectively, resolution) can at present only be captured from dedicated airborne platforms. It is not suitable for our applications due to the associated cost and logistic effort.

\section{Proposed Methodology}

\bk{}{
\begin{figure*}[!ht]
        \centering
        \includegraphics[width=0.75\linewidth]{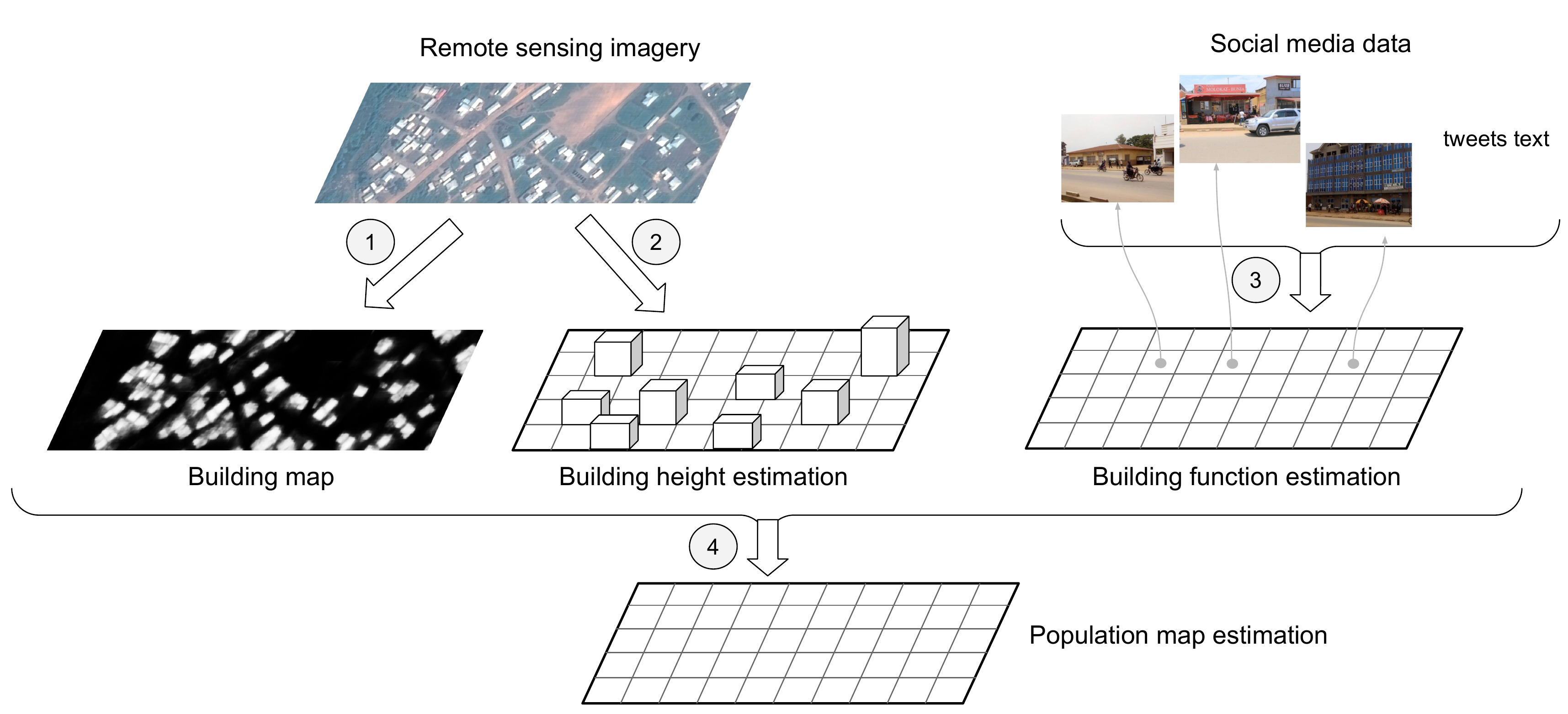}
        \caption{\bk{}{Overview of the data sources and methodologies. We intend to infer building footprints (1) and heights (2) from remote sensing imagery and building functions (3) from social media data, such as ground-based pictures. These intermediate results can then be fused (4) to obtain population density maps.}}
        \label{fig:flowchart-methodology}
\end{figure*}
}

\begin{figure*}[!ht]
     
    \begin{minipage}{0.15\linewidth}
        \centering
        \includegraphics[width=\linewidth]{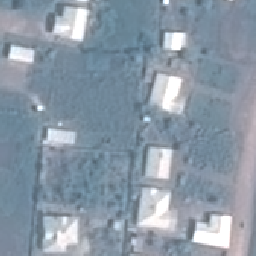}\\
        \vspace{2pt}
        \includegraphics[width=\linewidth]{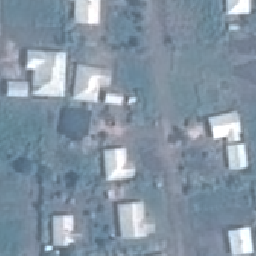}\\
    \end{minipage}
    ~
    \begin{minipage}{0.15\linewidth}
        \centering
        \includegraphics[width=\linewidth]{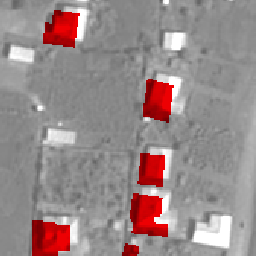}\\
        \vspace{2pt}
        \includegraphics[width=\linewidth]{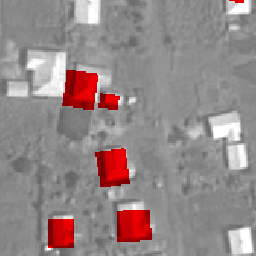}\\
    \end{minipage}
    ~
    \begin{minipage}{0.15\linewidth}
        \centering
        \includegraphics[width=\linewidth]{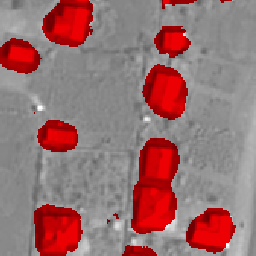}\\
        \vspace{2pt}
        \includegraphics[width=\linewidth]{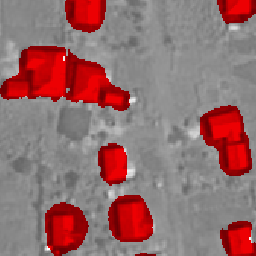}\\
    \end{minipage}
    ~
    \begin{minipage}{0.15\linewidth}
        \centering
        \includegraphics[width=\linewidth]{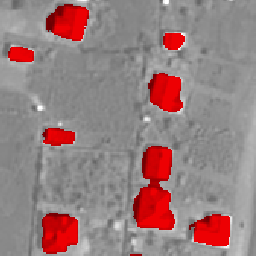}\\
        \vspace{2pt}
        \includegraphics[width=\linewidth]{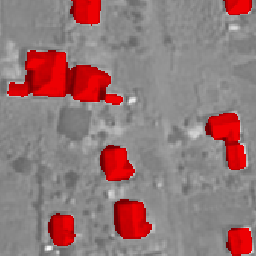}\\
    \end{minipage}
    ~
    \begin{minipage}{0.15\linewidth}
        \centering
        \includegraphics[width=\linewidth]{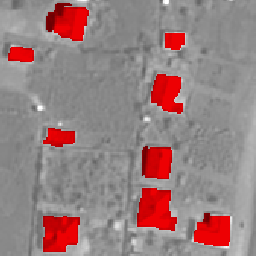}\\
        \vspace{2pt}
        \includegraphics[width=\linewidth]{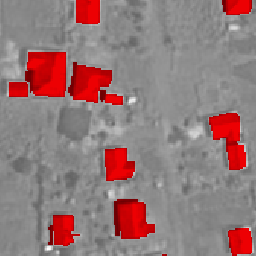}\\
    \end{minipage}
    
    \begin{minipage}{0.15\linewidth}
        \centering
        (a) Image
        \newline
    \end{minipage}
    ~
    \begin{minipage}{0.15\linewidth}
        \centering
        (b) Available reference data
    \end{minipage}
    ~
    \begin{minipage}{0.15\linewidth}
        \centering
        (c) Supervised result
        \newline
    \end{minipage}
    ~ 
    \begin{minipage}{0.15\linewidth}
        \centering
        (d) Weakly supervised result~\cite{daudt2019gad}
    \end{minipage}
    ~
    \begin{minipage}{0.15\linewidth}
        \centering
        (e) Post-processed result
    \end{minipage}
    
    \caption{\daudt{}{Preliminary building segmentation results in Bunia, obtained with a U-Net~\cite{ronneberger2015u} with EfficientNet~\cite{pmlr-v97-tan19a} backbone. The inaccurate building annotations in the available reference data lead to inaccurate prediction boundaries. This effect can be mitigated through the use of iterative label cleaning.}}
    \label{fig:prelim-results}
\end{figure*}

The goal of our project is to map populations and population densities with the help of satellite data and social media posts. \bk{In detail, this includes the following parts:}{Figure~\ref{fig:flowchart-methodology} shows the data sources and methodologies considered to obtain per-building estimates of inhabitants and thus population densities. The numbers highlight different methodological parts as follows:}
\begin{enumerate}
    \item In a first step, \bk{}{we} generate high-resolution building maps with the help of deep \bk{}{convolutional} neural networks (DNNs) \bk{}{such as U-Net~\cite{ronneberger2015u}}. Different data sources can be used to this end, including very high-resolution, commercial satellite imagery as well as elevation models. Besides providing a basis for planning in a more geometric sense, building footprints can serve as an initial approximation towards population density estimation. \john{}{Oftentimes, reference data to train a model are noisy, such as incompleteness issues as in the case of our building footprints ground truth (see Fig.~\ref{fig:prelim-results}(b)). To compensate for this problem, we will use a weakly supervised method based on~\cite{daudt2019gad} to generate building maps. Some preliminary results are depicted in Figure~\ref{fig:prelim-results}.}
    \item \bk{As an alternative, we will also investigate an approach that maps local population densities directly. Recent works on density estimation for agricultural~\cite{rodriguez2021rse} and environmental~\cite{bowler2019using} applications have shown great potential to directly regress densities from satellite images; we will evaluate that strategy for population density mapping.}{In a second step, we intend to estimate the height of buildings for a more precise estimation of the number of inhabitants. As outlined above, monocular height estimation is challenging and an open research question. Yet, multiple methods have been proposed and we plan on investigating some that use optical imagery directly~\cite{srivastava2017joint,mou2018im2height, 8891800}, or a combination of optical and SAR data~\cite{wegner2013combining}.}
    \item As explained above, precise population counts not only depend on building footprints, but also on the function of buildings and neighborhoods. While such information may be challenging to retrieve with satellite-based Earth observation, it can be extracted more easily from street-level imagery~\cite{srivastava2018land,srivastava2020fine} as well as social media content~\cite{alam2018twitter,imran2020using}: building attributes, such as store fronts and hospital signs, can be automatically detected in ground-level imagery and provide direct indicators. Moreover, demographic indicators such as age groups and social indicators can also contribute valuable knowledge, and can be extracted\bk{ -- in disaggregated and anonymized form --}{} from social media platforms~\cite{fatehkia2020mapping,fatehkia2020relative}. \bk{}{In this step we intend to train DNNs that classify building functions based on ground-level imagery, tweets, and place descriptions from these heterogeneous data sources.}
    \item Finally, all described intermediate products shall be combined to obtain more accurate and complete population maps. The hope is that this integrated product will combine the geometric precision of satellite-based Earth observation with the rich semantic and demographic detail of ground-level images and social media content. The fusion may be accomplished with another DNN that merges the heterogeneous inputs at the level of deep, latent features\bk{}{, e.g. using MCB~\cite{fukui_multimodal_2016} or MUTAN~\cite{ben_younes_mutan_2017},} and is trained to \bk{jointly predict multiple relevant outputs, such as building locations and heights, building functions, estimated number of people per building, and more}{spatially regress population density}.
\end{enumerate}

\bk{}{As a baseline, we will also investigate an approach that maps local population densities directly. Recent works on density estimation for agricultural~\cite{rodriguez2021rse} and environmental~\cite{bowler2019using} applications have shown great potential to directly regress densities from satellite images; we will evaluate that strategy for population density mapping.}

\bk{As an initial dataset for method development, w}{W}e target four representative cities and their surroundings, for which data are available: \bk{}{Gumbo (East Juba, South Sudan), Mokolo (Cameroon), Bunia (DRC), Tal Afar (Iraq). We note that all data (building footprints, tweets, ground-level imagery, etc.) are in disaggregated and anonymized form, without information on tenants or owners (building data) or possibilities of identifying the original owner or uploader (social media data).}

\section{\bk{}{Expected Challenges}}

\begin{figure*}[t]
     
    \begin{minipage}{0.18\linewidth}
        \centering
        \includegraphics[width=\linewidth]{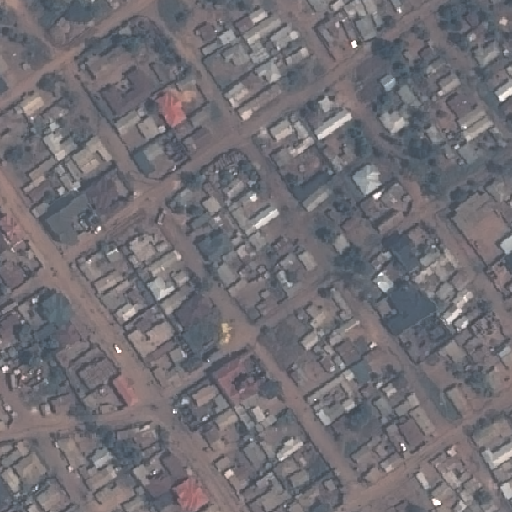}\\
        \vspace{5pt}
        \includegraphics[width=\linewidth]{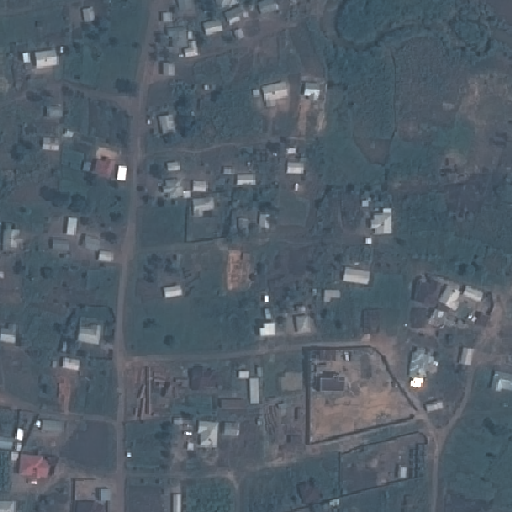}\\
        (a) Bunia
    \end{minipage}
    ~
    \begin{minipage}{0.18\linewidth}
        \centering
        \includegraphics[width=\linewidth]{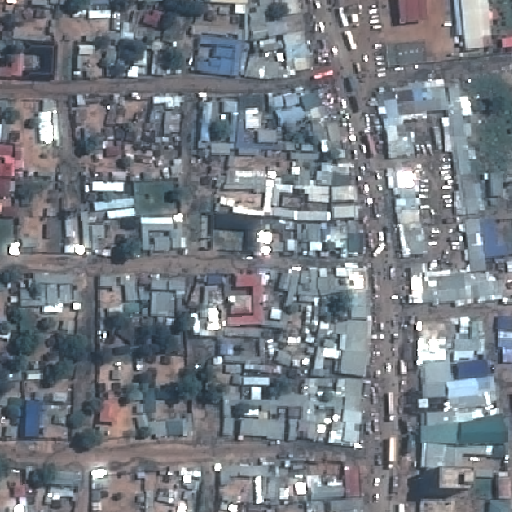}\\
        \vspace{5pt}
        \includegraphics[width=\linewidth]{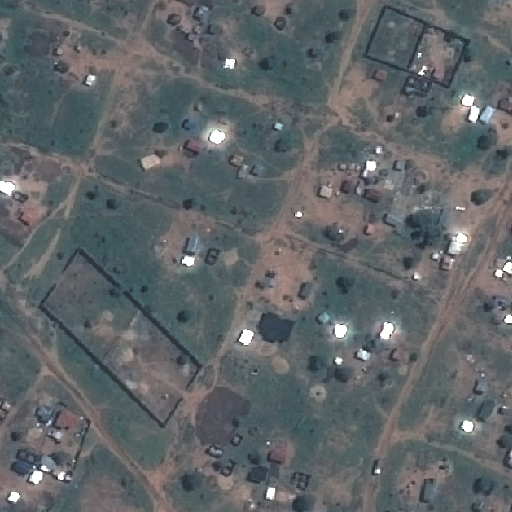}\\
        (b) Gumbo
    \end{minipage}
    ~
    \begin{minipage}{0.18\linewidth}
        \centering
        \includegraphics[width=\linewidth]{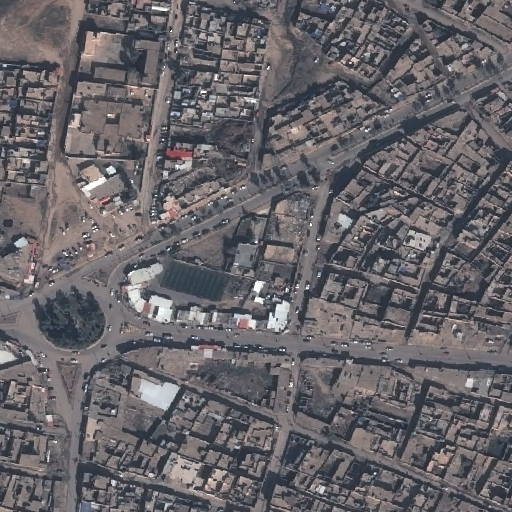}\\
        \vspace{5pt}
        \includegraphics[width=\linewidth]{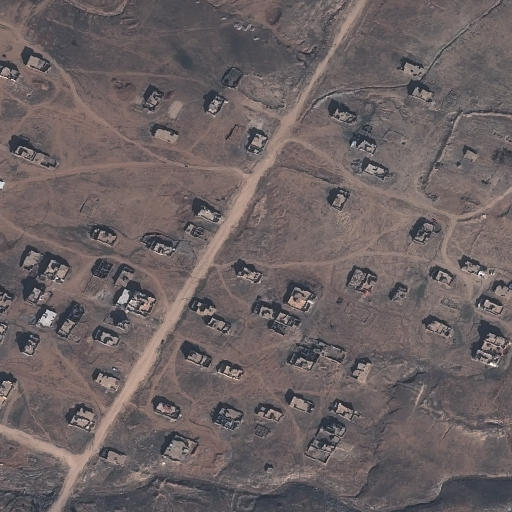}\\
        (c) \john{Talfar}{Tal Afar}
    \end{minipage}
    ~
    \begin{minipage}{0.18\linewidth}
        \centering
        \includegraphics[width=\linewidth]{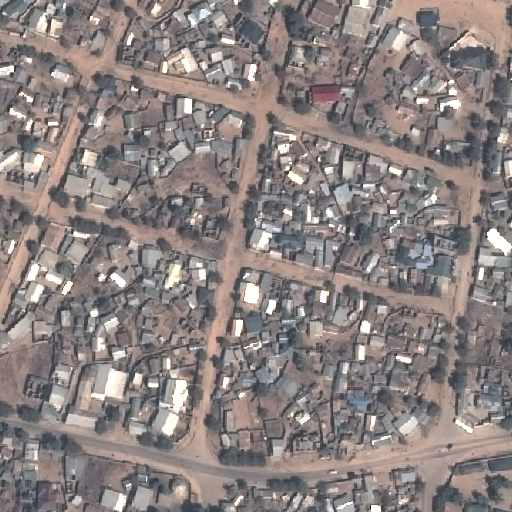}\\
        \vspace{5pt}
        \includegraphics[width=\linewidth]{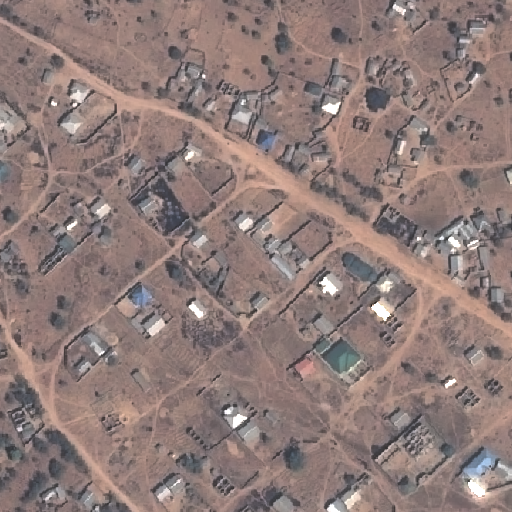}\\
        (d) Mokolo
    \end{minipage}
    \caption{The four regions initially exhibit considerable variations, such that semi-supervised or unsupervised methods for domain adaptation will play an important role to ensure generalization to target cities not seen during initial model training.}
    \label{fig:domain-shift}
\end{figure*}

Large variations between these four cities in terms of environmental conditions and structural lay-out give rise to significant domain shifts, which pose a particular challenge. See Figure~\ref{fig:domain-shift}. To ensure general applicability of the developed approaches, domain adaptation across geographic regions will therefore play a central role. We anticipate that this may include semi-supervised fine-tuning and possibly also unsupervised domain adaptation based on aggregate feature statistics, to limit the required amount of training labels for a new target region to a practically tractable magnitude~\cite{redko2019advances}.

\bk{}{Furthermore, there is a high probability that the coverage of social media data may not be sufficient, leaving us with incomplete information for some buildings. One possible means of compensation to this end may be to infer the building functions for some classes that are sufficiently distinguishable even from the air (e.g., private households versus large factory halls) and hence train a model to do so based on the satellite imagery.} \daudt{}{Nevertheless, partial information regarding building function could still be leveraged to improve population estimation results where available.}

\section{Overview}


The objective of this project is to develop methods that use heterogeneous sources of data to create fine-grained building maps and population distribution estimations for vulnerable regions in developing countries. We are hopeful that the methods (and in certain cases perhaps also the maps and population estimates) generated during the project could in the future support humanitarian organizations and make their mission planning more efficient.
We are excited about the opportunity to have geo-data experts from the ICRC collaborate directly with researchers in remote sensing, spatial data analysis and machine learning and hope that the joint effort between the humanitarian and academic organizations will be a catalyst to achieve our goals and help protect vulnerable populations.

\begin{acks}
This research is funded by the Science and Technology for Humanitarian Action Challenges (HAC) project.
\end{acks}

\bibliographystyle{ACM-Reference-Format}
\bibliography{main}

\appendix


\end{document}